\documentclass[runningheads]{llncs}
\usepackage{graphicx}
\usepackage{comment}
\usepackage{amsmath,amssymb} 
\usepackage{color}

\usepackage{epsfig}
\usepackage{stackengine}
\usepackage{url}            
\usepackage{booktabs}       
\usepackage{hyperref}       
\usepackage{tabu}           
\usepackage{multirow}
\usepackage{subcaption}
\usepackage{bigstrut}
\usepackage{graphicx}
\usepackage{amsmath}
\usepackage{algorithm}
\usepackage{algorithmicx}
\usepackage{algpseudocode}
\usepackage[dvipsnames]{xcolor}
\usepackage{stackengine}

\DeclareMathOperator*{\argmin}{arg\,min}

\begin{document}
\pagestyle{headings}
\mainmatter
\def\ECCVSubNumber{2255}  

\title{MetaDistiller: Network Self-Boosting via Meta-Learned Top-Down Distillation} 

\titlerunning{Meta-Learned Top-Down Self Distillation}
%
\author{Benlin Liu\inst{1}\and
Yongming Rao\inst{2}\and
Jiwen Lu\inst{2}\and
Jie Zhou\inst{2}\and
Cho-Jui Hsieh\inst{2}}
\authorrunning{Liu et al.}
%
\institute{University of California, Los Angeles \and
Tsinghua University \\
\email{\{liubenlin,chohsieh\}@cs.ucla.edu}, \email{raoyongming95@gmail.com}, \email{\{lujiwen, jzhou\}@tsinghua.edu.cn}}
\maketitle

\begin{abstract}
Knowledge Distillation (KD) has been one of the most popular methods to learn a compact model. However, it still suffers from high demand in time and computational resources caused by sequential training pipeline. Furthermore, the soft targets from deeper models do not often serve as good cues for the shallower models due to the gap of compatibility. In this work, we consider these two problems at the same time. Specifically, we propose that better soft targets with higher compatibility can be generated by using a label generator to fuse the feature maps from deeper stages in a top-down manner, and we can employ the meta-learning technique to optimize this label generator. Utilizing the soft targets learned from the intermediate feature maps of the model, we can achieve better self-boosting of the network in comparison with the state-of-the-art. The experiments are conducted on two standard classification benchmarks, namely CIFAR-100 and ILSVRC2012. We test various network architectures to show the generalizability of our MetaDistiller. The experiments results on two datasets strongly demonstrate the effectiveness of our method. 

\keywords{Knowledge Distillation, Meta Learning}
\end{abstract}

\section{Introduction}

\begin{figure}[htb]
  \centering
  \includegraphics[width=\textwidth]{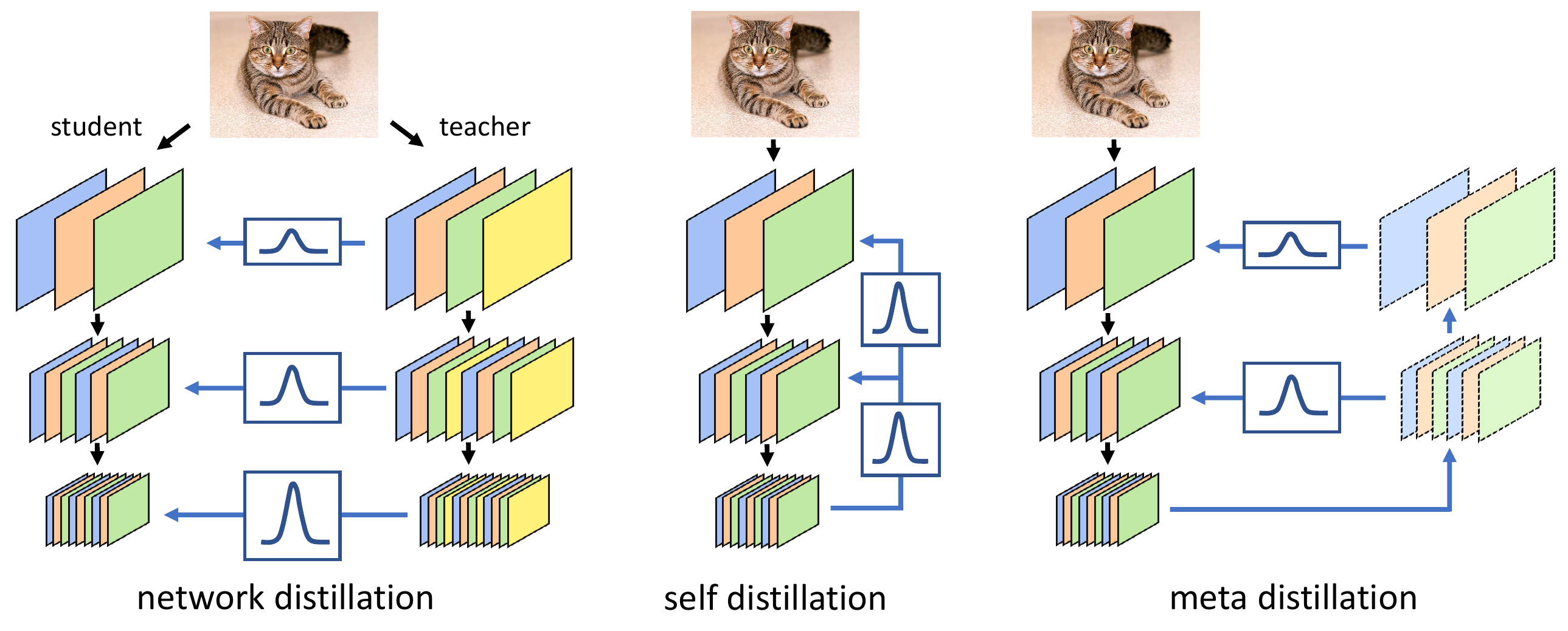}
  \caption{\small \textbf{The key idea of our approach.} Traditional knowledge distillation methods often follow the time-consuming teacher-student pipeline as left. To overcome this problem, self-distillation (middle) abandons the large teacher model and uses the final output as soft teacher target for all intermediate outputs, which may be hampered by the capacity gap between deep layers and shallow layers~\cite{cho2019efficacy}. Our MetaDistiller (right) proposes to generate more compatible soft target for each intermediate output respectively in a top-down manner.Best viewed in color.}
  \label{fig:intuition }
\end{figure}

Although deep neural networks have achieved astonishing performance in many important computer vision tasks, they usually require  large number of model parameters (weights) which lead to extremely high time and space complexity in training and deployment. 
As a result, there exists a trade-off between accuracy and efficiency. 
To have a small and efficient model with similar performance as a deep one, 
many techniques have been proposed,  including pruning~\cite{han2015deep,lin2017runtime,rao2018runtime}, quantization~\cite{han2015deep,polino2018model}, low-rank decomposition~\cite{yu2017compressing,chen2018groupreduce} and any others. Knowledge Distillation (KD)~\cite{hinton2015distilling} is one of the most popular methods among them to train a compact network with high performance. Specifically, it first trains a teacher model, then uses the output of this teacher model as an additional soft target to transfer the knowledge learned by the teacher to the student model. This distillation process may go on for even more than one generation. Many variants~\cite{bagherinezhad2018label,furlanello2018born,romero2014fitnets,zagoruyko2016paying,zhang2018deep} have been proposed subsequently to improve the guidance from teacher to student.

However, all these conventional distillation methods perform in a sequential way which is difficult to parallelize. This indicates that the distillation process may take a lot of time, especially for the learning of teacher model, while our goal is the compact student model. Hence, some works have explored to sidestep this sequential procedure.~\cite{yang2019snapshot} proposes to use the output of previous iteration as the soft target for the current iteration, but this may lead to amplified errors, and it's rather hard to choose which previous iteration to use. Besides, either the generated soft targets of the whole dataset or the model at a certain iteration has to be kept in memory since they will serve as teachers. \cite{zhang2019your} proposes a self distillation method in the complete sense by extending~\cite{lee2015deeply}, which adds classifiers to the intermediate hidden layers to directly supervise the learning of shallower layers.~\cite{zhang2019your} further adopts the output of the final layer as teacher to transfer the knowledge from deep layers to the shallow ones. However, this design still suffers from the problem  pointed out by~\cite{cho2019efficacy} as conventional distillation methods.~\cite{cho2019efficacy} finds that larger models do not often make better teachers due to the mismatched capacity between large models and smaller ones, which makes small students unable to mimic the large teachers. We can observe such problem in self-distillation given the capacity gap between the deepest model and the shallower ones.

To overcome these problems in self-distillation, we propose a new method called MetaDistiller where a label generator is employed to learn the soft targets for each shallower output layers instead of using the output of the last layer. We expect the learned soft targets are more matched with the capacity of the corresponding shallower output layers in the model and our model can thus be better self-boosted by using these learned soft targets. Our method also does not need a teacher model and therefore avoids the time-consuming sequential training. The soft targets for shallower output layers are obtained based on the feature maps from deeper layers, which is referred to as \textbf{Top-Down Distillation}. This is inspired by the observation that the feature maps in deep layers often encode higher-level and more task-specific information compared to the ones in shallower layers, and the soft targets should transfer these knowledge to the shallower layers for better training of the model. We argue this shares the same principle with the retrospection of human beings given that we can be aware of what we have missed in the previous stage as the learning proceeds. The top-down distillation aims at correcting this defect. Moreover, unlike~\cite{romero2014fitnets}, we learn the soft targets for the output probability distribution from intermediate layers rather than for the feature maps since the categorical information is shared across different stages while the contextual information needed for the reconstruction is not. To generate the soft targets that are more compatible with the intermediate outputs, the label generator is optimized by bi-level optimization which is commonly used in meta-learning, 
given we share the same motivation of `learning to learn'. Though additional computational burden is added during training in comparison with self-distillation due to the existence of meta learning, it is relatively marginal when compared to traditional teacher-student distillation methods. Besides, neither time nor space complexity increases during inference time since label generator is then abandoned. Hence, such trade-off for the improvement in compatibility of soft targets is quite cost-effective.Fig.~\ref{fig:intuition } illustrates our motivation intuitively.

To show the effectiveness of our approach, we conduct experiments for various network architectures on two standard classification benchmarks, namely CIFAR-100~\cite{krizhevsky2009learning} and ILSVRC2012~\cite{ILSVRC15}. The empirical study shows that we can consistently attain better results by employing our MetaDistiller compared to both self-distillation and traditional teacher-student distillation methods. We also find that the generated soft targets are more complementary to the one-hot ground-truth label compared to the final output. This is because the generated soft targets can better capture the information among negative classes, which is not covered by one-hot ground-truth, and thus can serve as a better regularization term to help the model generalize well to unseen data.

\section{Related Work}
\noindent \textbf{Knowledge Distillation}
Knowledge distillation is widely adopted to train a shallow network to achieve comparable or higher performance than a deep network. Hinton et al.~\cite{hinton2015distilling} first proposes the idea of knowledge distillation where a well-optimized large model is employed as teacher to provide additional information not contained in the one-hot label to the compact student network. Fitnets~\cite{romero2014fitnets} improves upon~\cite{hinton2015distilling} by using the feature maps in the intermediate layers of teacher net as hints.~\cite{zagoruyko2016paying}
trains the student model to mimic the attention maps from the teacher model instead.~\cite{zhang2018deep} blurs the difference between teacher and student by exchanging their roles iteratively, which is similar to co-training. However, all these traditional distillation methods perform the time-consuming sequential training. Although~\cite{furlanello2018born,bagherinezhad2018label} share the same architecture between teacher and student, they still require sequential training since each student is trained from scratch to match the teacher's soft labels.

To sidestep this resource-consuming pipeline, some works~\cite{yang2019snapshot,zhang2019your} explore to distill the network itself.~\cite{yang2019snapshot} uses the outputs from previous iteration as soft targets, but this risks amplifying the error in the learning process, and decision on which iteration to adopt as teacher is rather hard. Strictly speaking, it is not self-boosting in a complete sense as we still need to keep the model of certain iteration as teacher.~\cite{zhang2019your} improves upon~\cite{lee2015deeply} by further employing the final output as teacher to supervise the output from intermediate layers. Their self-distillation method totally abandons the need of keeping a big teacher model or the soft targets of the whole dataset produced by it to perform the distillation. Nevertheless, this simple formulation suffers from the problem as illustrated in~\cite{cho2019efficacy} that shallower student is difficult to directly mimic the output from deep teacher model due to the gap between capacity. Our method overcomes this problem while still performs the distillation without resource-consuming teacher-student pipeline by generating more compatible soft targets through a meta-learned label generator to transfer the knowledge in a top-down manner.

\noindent \textbf{Meta Learning}
The core idea of meta learning is `learning to learn', which means taking the optimization process of learning algorithm into the consideration of optimizing the learning algorithm itself. Bengio et al.~\cite{bengio2000gradient} first applies meta learning to optimize hyper-parameters by differentiating through the preserved computational graph of training phase to obtain the gradients with regard to the hyper-parameters. Finn et al.~\cite{finn2017model} uses meta learning to learn better initialization for few-shot learning.~\cite{wang2018low} utilizes meta learning to augment training data for low-shot learning.~\cite{jang2019learning} incorporates meta learning into traditional knowledge distillation. They automatically decide what knowledge in the teacher as well as its amount will be transferred to which part of the student based on meta gradients instead of hand-crafted rules. Sharing the similar motivation with the works mentioned above, in this paper, we adopt meta learning to learn the soft targets for self knowledge distillation, which can be subsequently used as supervision signals to boost the learning of the model. 

\section{Approach}
In this section, we will introduce the proposed method, MetaDistiller. 
We first briefly review the formulation of knowledge distillation in Section~\ref{sec:formulation}, and then extend it to self-boosting in Section~\ref{sec:self}.
\begin{figure}[htb]
  \centering
  \includegraphics[width=\textwidth]{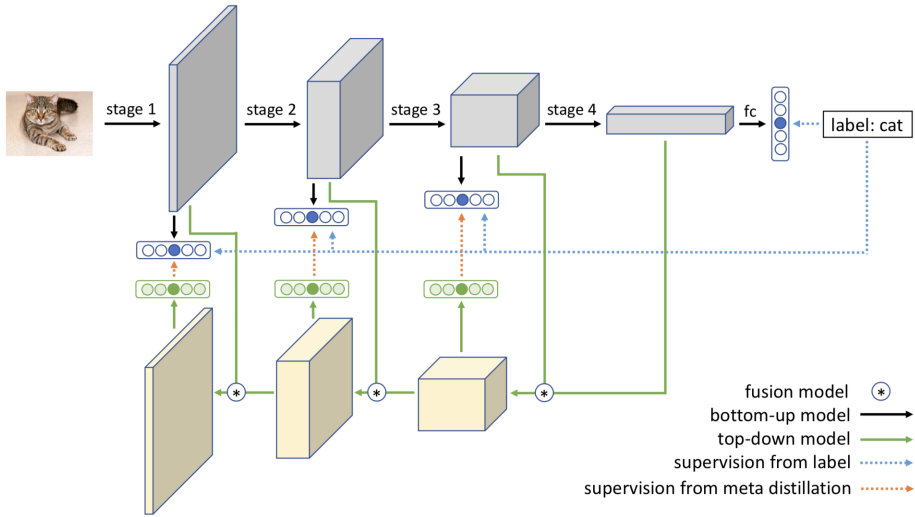}
  \caption{\small \textbf{The pipeline of our approach.} The grey part and the yellow part correspond to main model and label generator respectively. We fuse the feature map produced in the feed forward process of main model through a top-down manner by adopting the operation defined in Equation~\eqref{eq:fusion} iteratively to generate soft teacher targets (green). The intermediate outputs (blue) of the main model are supervised by the ground-truth label and generated soft teacher targets simultaneously. The label generator is trained by using meta learning. Best viewed in color.}
  \label{fig:model}
\end{figure}
In Section~\ref{sec:top-down}, we propose to perform the top-down distillation by incorporating feature maps from different stages progressively to generate soft targets. In Section~\ref{sec:meta}, we then discuss how to apply meta learning to learn the soft target generator. Finally, we will present the whole training and inference pipeline of the proposed algorithm  in Section~\ref{sec:training}.

\subsection{Background: Knowledge Distillation}\label{sec:formulation}
Knowledge Distillation (KD), proposed by Hinton et al.~\cite{hinton2015distilling}, aims at transferring knowledge from a deep \textbf{\textit{teacher}} network, denoting as \textit{T}, to a shallow \textbf{\textit{student}} network, denoting as \textit{S}. In order to  transfer knowledge from teacher to student, the loss for training student network  is modified by adding KL divergence
between the teacher's and the student's  output  probability distributions. 
Formally, given labeled dataset $\mathcal{D}$ of $N$ samples $\mathcal{D} = \{\left(x_1, y_1\right), \dots, \left(x_N, y_N\right)\}$, we can write the loss function of student network as following,
\begin{equation}
    \mathcal{L}_{S}\left(\mathcal{D};  \theta_{S} \right)= \frac{1}{N} \sum_{i=1}^N \alpha\mathcal{L}_{CE}\left(y_i,  \textit{S}\left(x_i; \theta_{S} \right) \right) + \left(1 - \alpha\right) {\tau^2} \mathcal{KL}\left(\textit{T}\left(x_i; \theta_{T}\right) || \textit{S}\left(x_i; \theta_{S} \right)\right),  \label{eq:loss}
\end{equation}
where $\alpha$ is the hyper-parameter to control the relative importance of the two terms, $\tau$ is the temperature hyper-parameter,  $\theta_{T}$ and $\theta_{S}$ are the parameters of teacher $\textit{T}$ and student $\textit{S}$ respectively. $\mathcal{L}_{CE}$ refers to  the cross entropy loss and $\mathcal{KL}$ refers to the KL divergence which measures how one probability distribution is different from another. 

By minimizing this modified loss function, the student network will try to match 
the one-hot ground-truth distribution (the first term) while lowering its discrepancy with 
output probability distribution from teacher $\textit{T}\left(x_i; \theta_{T} \right)$ to transfer the knowledge of negative classes learned by teacher network. The probability distribution over classes can be obtained by employing softmax function, where the probability of sample $s_i$ belonging to class $j$ can be expressed as,
\begin{equation}
     P\left(y_i = j | x_i \right) = \frac{e^{{z^j_i}/\tau}}{\sum_{c=1}^{\mathcal{C}} e^{z^c_i / \tau}}
\end{equation}
where $\mathcal{C}$ is number of classes, $z_i$ is the output logits of sample $x_i$, $\tau$ is the same temperature hyper-parameter as~\eqref{eq:loss}. The role of $\tau$ is to control the smoothness of the output probability distribution. The higher the temperature, the higher the entropy of distribution, i.e. the smoother distribution.

\subsection{Network Self-Boosting}\label{sec:self}
One of the major problems of the above teacher-student pipeline is the long training time caused by the two-stage sequential training procedure. 
To overcome such difficulty, a method called self-distillation~\cite{zhang2019your} has been proposed by adopting deep-supervised net~\cite{lee2015deeply} to achieve self-boosting without first training a large teacher network. 
In the following, we will introduce a general self-boosting framework where \cite{zhang2019your} is a special case under our framework, and then we will show our general framework leads to improved algorithms in the next two subsections. 


In the self-boosting framework introduced in \cite{lee2015deeply}, 
the deep network is divided into $K$ stages and there is an exit branch at the end of each early stage.  In addition to the loss at the final output, the one-hot ground-truth label is also used to supervise these early stages. Besides using the one-hot hard label as supervision signal to the intermediate outputs, we propose to add soft teacher targets to supervise these intermediate outputs. 
Specifically, \eqref{eq:loss} can be modified as following: 
\begin{equation}
    \mathcal{L}_{S}(\mathcal{D};  \theta_{S}, T_k)= \frac{1}{N} \sum_{i=1}^N
    {\bigg(} \mathcal{L}_{CE}\left(y_i,  \textit{S}_{K}\left(x_i; \theta_{S} \right) \right) +  \sum_{k=1}^{K-1} \mathcal{L}_{k}(\textit{S}_k\left(x_i; \theta_{S} \right), y_i; T_k)
    {\bigg)}
    \label{eq:self_boost_loss}
\end{equation}
where $\textit{S}_{k}$ indicates the output of the $k$-th stage in the student network, and $\textit{S}_{K}$ is the final output of the model. $T_k$ is called the {\bf soft teacher target} for stage $k$ to perform  knowledge distillation, which is a function depends on stage $k+1, \dots, K$ of the network. 
Unlike~\eqref{eq:loss}, there is no distillation term for the final output $\textit{S}_{K}$ since we do not have a teacher model to provide such supervision signal now. $\mathcal{L}_k$ refers to the loss function of outputs at earlier stages. For the first $K-1$ stages, the supervision signal of stage $k$ can be formally written as
\begin{equation}
    \mathcal{L}_{k}(\textit{S}_k(x_i; \theta_{S}), y_i; T_k) = \alpha \mathcal{L}_{CE}\left(y_i,  \textit{S}_{k}\left(x_i; \theta_{S} \right) \right) + \left(1 - \alpha\right) {\tau^2} \mathcal{KL}\left(T_k || \textit{S}_{k}\left(x_i; \theta_{S} \right)\right) \label{eq:stage_loss}
\end{equation}
where outputs from the intermediate layers are supervised by hard labels and soft teacher targets simultaneously.

In self-distillation~\cite{zhang2019your}, they directly utilize the output from the last layer $\textit{S}_{K}\left(x_i; \theta_{S} \right)$ as $T_k$. 
However, we will argue that this is not the best way to assign $T_k$ and will propose an automatic way to generate $T_k$ that can significantly improve the performance. 

\subsection{Top-Down Distillation}\label{sec:top-down}

As mentioned in the previous subsection, self-distillation~\cite{zhang2019your} sets the soft teacher target $T_k$ as the final layer softmax output for all the intermediate stages, and we argue this is not the best way for self-supervision. 
First, there exists ``gap of capacity'' problem, which means the capacity of earlier stages may not be enough to learn the final softmax output of a deep network. Furthermore, compared to the output probability distribution, feature maps in the deeper layers undoubtedly contain more information beyond the categorical information,  which shall be useful for getting better soft teacher targets for earlier stages.


Based on these ideas, we propose to distill the knowledge from the feature maps of deep layers to better supervise the learning of shallower layers through a top-down feature fusion process. This fusion process can be recursively defined as,
\begin{equation}
    \mathcal{\hat{F}}_k = {\rm conv_{3\times3}}({\rm UpSampling_{\times2}}(\mathcal{\hat{F}}_{k+1}) +  {\rm conv_{1\times1}}(\mathcal{F}_k))\label{eq:fusion}
\end{equation}
where $\mathcal{F}_k$ indicates the original output feature map of the $k$-th stage in the main model and $\mathcal{\hat{F}}_k$ indicates the feature map produced by our label generator based on feature maps of stage $k+1, \dots, K$. For the final stage $K$, we slightly modify \eqref{eq:fusion} by only keeping the $1\times1$ convolution operation since we do not have $\mathcal{\hat{F}}_{K+1}$. Upsampling can be simply achieved by using bilinear interpolation to match the spatial size of the the feature map from two neighboring stages as each stage scales the input feature map by 0.5. $1\times1$ convolution is responsible for matching the channel size to perform element-wise addition for fusion. We finally use a $3\times3$ convolution layer to refine this result to obtain the generated feature map $\mathcal{\hat{F}}_k$ for the $k$-th stage.
Note that the parameters for these convolution kernels are learnable which makes our approach able to automatically generate the best soft target feature for each particular task. We will  discuss how to learn these parameters in the next subsection. 

The process described above can be considered as the retrospection of human beings if we take the feed forward of neural network as the learning process of humans. Through rethinking what we have missed in previous learning stages, we could correct this for better learning if we could access the past states. And the top-down distillation aims at transferring such retrospection back to early stages for better learning. Compared to categorical probability distribution, feature maps contain more information, thus can serve a better indicator for different learning stages to achieve a better retrospection.

Though we can directly supervise the discrepancy between $\mathcal{F}_k$ and $\mathcal{\hat{F}}_k$ by using loss function like $L_2$ distance, we argue this is not a good choice based on the fact that feature also encodes contextual information other than categorical information. Unlike categorical information which should be invariant across different stages, the downsampling operation often causes a loss in the contextual information. As a result, the evolution of feature maps is not fully invertible. Hence, instead of adding supervision signal on the feature map, we prefer to only supervise categorical information as the deep layers can help on this. We therefore learn a normalized probability distribution from $\mathcal{\hat{F}}_k$ as our $T_k$ by using an additional bottleneck module, which is also employed to produce the intermediate outputs, 
and use $T_k$ as soft teacher label to perform the knowledge distillation. The whole pipeline is illustrated in Fig.~\ref{fig:model}

\subsection{Meta-Learned Soft Teacher Label Generator}\label{sec:meta}
The soft teacher target generator defined in \eqref{eq:fusion} is actually a fusion network that can subsequently help us to learn a better model, and therefore learning the parameters of this generator network shares the same  ``learning-to-learn'' intuition with meta learning. In the following, we show how to learn the label generator based on the commonly used bi-level optimization framework in meta learning. 

Formally, denoting the parameters of the main network as $\theta_s$ and the parameters of label generator as $\theta_g$, learning the label generator can be written as a bi-level problem, 
\begin{eqnarray} \nonumber 
      && \min_{\theta_g} \ \mathcal{L}^{\rm val} (X_{\rm val}, Y_{\rm val} ; \theta_s^*) \\ 
      s.t. && \theta_s^*  = \argmin_{\theta_s} \ \mathcal{L}_{S}(\mathcal{D_{\rm train}};  \theta_{s}.  T_k(\theta_g)),  \label{eq:optimization}
\end{eqnarray}
where $\mathcal{L}^{\rm val}$ is the standard cross entropy loss and $\mathcal{L}^{\rm train}  (X_{\rm train}, Y_{\rm train} ; \theta_s, \theta_g)$ is the same loss function as~\eqref{eq:self_boost_loss} only with $T_k$ changing to $T_k(\theta_g)$ as the soft teacher targets in our framework is generated through a label generator parameterized by $\theta_g$. 

To conduct updates on $\theta_g$, we need to compute  $\nabla_{\theta_g} \mathcal{L}^{\rm val} (X_{\rm val}, Y_{\rm val} ; \theta_s^*)$. 
However this is nontrivial since $\theta_s^*$ is an implicit function of $\theta_g$. This could be potentially done by the implicit function theory but the exact solution will lead to huge computational overhead. Therefore, following a standard approach used in meta learning (e.g., \cite{finn2017model}), 
%
we only approximate the $\theta_s^*$ by the current one with an additional gradient update step. This approximation can be written as
\begin{equation}
    \theta_s^+ := \theta_s - \zeta \nabla_{\theta_s}
    \mathcal{L}_{S}(\mathcal{D_{\rm train}};  \theta_{s}.  T_k(\theta_g)),
    \label{eq:inner}
\end{equation}{}
where $\zeta$ is the learning rate for this inner optimization step.

According to the above procedure, gradient with respect to $\theta_g$ can be expressed formally as
\begin{eqnarray}
      && \nabla_g \ := \nabla_g \mathcal{L}^{\rm test} (X_{\rm test}, Y_{\rm test} ; \theta_s^+) \nonumber\\
      s.t. && \theta_s^+ \ = {\rm opt}_{\theta_g}(\theta_s - \zeta \nabla_{\theta_s}\mathcal{L}_{\rm train}(X_{\rm train}, Y_{\rm train} ; \theta_s, \theta_g)),  \label{eq:theta_g}
\end{eqnarray}
where ${\rm opt}_{\theta_g}(\cdot)$ indicates the term inside the parentheses is differentiable with regard to $\theta_g$ for optimization. 

\renewcommand{\algorithmicrequire}{\textbf{Input:}}
\renewcommand{\algorithmicensure}{\textbf{Output:}}
\begin{algorithm}[htb] \small
   \caption{Training process for solving optimization problem in~(\eqref{eq:optimization}):}
   \label{alg:train}
   \begin{algorithmic}[1] 
      \Require labeled dataset $\mathcal{D} = \{(x_i, y_i)\}$, main model $\textit{S}(\theta_s)$, label generator $\textit{G}(\theta_g)$
      \Ensure Model $\textit{S}^{*}(\theta_s)$
      \State \textbf{initialize:} main model $\textit{S}(\theta_s)$, label generator $\textit{G}(\theta_g)$
      
      \While{not converged}
      \State // train $\textit{S}$
      \State Sample a random mini-batch $\{(x_i, y_i)\}$ from $\mathcal{D}$
      \State Forward pass on  
      $S$ to obtain $K$ outputs $\{\textit{S}_1(x_i; \theta_s), \cdots, \textit{S}_K(x_i; \theta_s) \}$
      \State Forward pass on $G$ 
      to generate $\{T_1, \cdots, T_{K-1}\}$
      \State Optimize $\textit{S}$ following~\eqref{eq:self_boost_loss} using generated soft targets $\{T_1, \cdots, T_{K-1}\}$
      \State // train $\textit{G}$ for every $\rm M$ epochs
      \If{epoch\_num \% $\rm M$ ==0}
        \State Randomly split $\mathcal{D}$ in half to get two disjoint dataset $\mathcal{D}_{\rm train}$ and $\mathcal{D}_{\rm test}$
        \State //inner update, retaining the computational graph
        \State sample a random mini-batch $X_{\rm train}$ with labels $Y_{\rm train}$ from $\mathcal{D}_{\rm train}$
        \State compute $\theta_s^+$ by using a single step gradient update following~\eqref{eq:inner}
        \State //meta gradient step
        \State sample a random mini-batch $X_{\rm test}$ with labels $Y_{\rm test}$ from $\mathcal{D}_{\rm test}$
        \State Calculate $\mathcal{L}_{\rm test}(\theta_s^+)$ using cross entropy loss
        \State Obtain $\nabla_g$ following~\eqref{eq:theta_g}
        \State //update label generator
        \State Update $\theta_g$ based on $\nabla_g$ using gradient descent
      \EndIf
      \EndWhile
   \end{algorithmic}
\end{algorithm}

\subsection{Training and Inference}\label{sec:training}

The whole training pipeline of our method is described in Algorithm.~\ref{alg:train}. More specifically, we alternatively update $\theta_s$ (model parameters) and $\theta_g$ (generator's parameters) by SGD, where the gradient with respect to generator is computed by \eqref{eq:theta_g}. The generator is used to perform top-down distillation to generate the soft teacher targets to facilitate the self-boosting of the main model. 
The reason why we split the whole training dataset into two disjoint sets of equal size is that we expect the model supervised by generated soft targets can generalize well to unseen data, which can be taken as an indicator of higher compatibility. We set $\rm{M} = 5$, i.e. we update the label generator every 5 epochs. 
This can reduce the training time while still keep the performance from degradation. 
For instance, on ResNet18~\cite{he2016deep}, the training time increases by 1.1h compared to simply training the model without any types of distillation while it takes 7.9h to first train a ResNet101~\cite{he2016deep} as teacher model if we perform the conventional two-stage knowledge distillation. Moreover, the performance of the well-trained ResNet101 is almost the same as our self-boosted ResNet18. This shows that our proposed method can preserve the speed of self-distillation while improving its final test accuracy.

During the inference phase, we can abandon the label generator and the layers utilized to produce the intermediate results, so there will be no additional computational cost required for deployment compared to the model trained by simply using the cross entropy loss. Furthermore, if the computational resources allow, we can further perform ensemble by using the intermediate output. We report the results of both the final output and the ensemble output in experiments, and ensemble often improves the accuracy of final output by around $1\%$.
and ensemble often improves the accuracy of final output by around $1\%$.
\section{Experimental Results}



\subsection{Setups}
We evaluate our method on two widely used image classification datasets, namely CIFAR-100~\cite{krizhevsky2009learning} and ILSVRC2012~\cite{deng2009imagenet}. To verify the generalizability of our method, we test with multiple different network architectures. As for hyper-parameters, we set balance weight $\alpha$ to 0.5 and temperature $\tau$ simply to 1 for all the datasets constantly. For optimization, we use SGD and Adam~\cite{kingma2014adam} optimizer for main model $\textit{S}$ and label generator $\textit{G}$ respectively with initial learning rates both as 0.1. We train the network for 200 epochs in total, and multiply the learning rate by 0.1 at epoch 80 and 140. The learning rate of inner step gradient update $\zeta$ is kept the same with the learning rate of main model $\textit{S}$. All the experiments are conducted with PyTorch. 

\subsection{CIFAR-100}
\begin{table*}[!tbp]
\centering
\caption{\small \textbf{Our method consistently outperforms the the self-distillation on CIFAR-100.} We present the results on CIFAR-100 by testing our MetaDistiller on 8 different networks. We report the result in~\cite{zhang2019your} in the parentheses.}
\label{tab:CIFAR} 

\centering
    \begin{tabu}to\textwidth{l*{6}{X[c]}}\\\toprule
    \multirow{2}{*}{Model} & \multirow{2}{*}{baseline} &  \multicolumn{2}{c}{Final Output} & \multicolumn{2}{c}{Ensemble Output} \\
    
    &         & \footnotesize{SD} & \footnotesize{MD} & \footnotesize{SD} & \footnotesize{MD} \\
    \midrule
  ResNet18\cite{he2016deep} &77.31(77.09) & 78.25(78.64) & \textbf{79.05} & 79.32(79.67) & \textbf{80.05}  \\
  ResNet50\cite{he2016deep} &77.78(77.68) & 79.71(80.56) & \textbf{80.85} & 80.38(81.04) &  \textbf{81.41} \\
  ResNet101\cite{he2016deep} &78.92(77.98) & 80.92(81.23) & \textbf{81.39} & 81.55(82.03) & \textbf{81.98} \\
  ResNeXT29-2\cite{xie2017aggregated} &77.78 & 79.43 & \textbf{80.38} & 80.14 & \textbf{80.89}   \\
  WideResNet16-1\cite{zagoruyko2016wide} &67.58 & 69.42 & \textbf{70.30} & 70.17 &  \textbf{70.81} \\
  VGG19(BN)\cite{simonyan2014very} &71.61(64.47) &74.27(67.73) & \textbf{75.12} &75.01(68.54) & \textbf{75.93} \\
  MobileNetv2\cite{sandler2018mobilenetv2} &71.81 & 74.46 & \textbf{75.37} & 75.33 & \textbf{75.96} \\
  Shufflenetv2\cite{ma2018shufflenet}&68.12 & 69.89 & \textbf{70.66} & 70.71 & \textbf{71.29}    \\
    
    \bottomrule
\end{tabu}
\end{table*}
We first conduct experiments on CIFAR-100~\cite{krizhevsky2009learning} for all 8 architectures mentioned in Sec 4.2. For all three different types of ResNets~\cite{he2016deep} and MobileNetv2~\cite{sandler2018mobilenetv2}, we divide the network into four stages, and the output feature maps for each of the first three stages are taken as input to a bottleneck module, following self-distillation~\cite{zhang2019your}, to produce the intermediate output for that stage. For ResNeXt~\cite{xie2017aggregated}, WideResNet~\cite{zagoruyko2016wide} and Shufflenetv2~\cite{ma2018shufflenet}, the network is divided into three stages and use the output feature maps from first two stages to produce intermediate output in the similar way. For VGG19~\cite{simonyan2014very}, we directly perform the average pooling and use a fully connected layer to produce the intermediate output for the feature maps before the second, third and fourth max-pooling operation. We implement the experiments on all other network architectures for self-distillation and use it as our baseline results. We also report the numbers in their original paper regarding three ResNet-type model and VGG. In Table.~\ref{tab:CIFAR}, we denote self-distillation as SD and denote our MetaDistiller as MD.

The detailed experimental results are listed in Table.~\ref{tab:CIFAR}. Besides the accuracy of final output, we also report the performance of the ensemble of all outputs. We can observe that our MetaDistiller consistently outperforms the baseline and self-distillation, indicating the generated soft teacher targets can better facilitate the training compared to the final output, and this generalize to different network architectures. The average improvement for three-stage model is slightly smaller due to less supervision is injected into the training of early stage.

\subsection{ILSVRC2012}
\begin{table*}[!tbp]
\centering
\caption{\small \textbf{Our method consitently outperforms the self-distillation on ImageNet.} We present the results on ImageNet by testing our MetaDistiller on 3 different networks. }
\label{tab:ImageNet} 
\centering
    \begin{tabu}to\textwidth{l*{5}{X[c]}}\\\toprule
    \multirow{2}{*}{Model} & \multirow{2}{*}{baseline} &  \multicolumn{2}{c}{Final Output} & \multicolumn{2}{c}{Ensemble Output} \\
    
    &         & \footnotesize{SD} & \footnotesize{MD} & \footnotesize{SD} & \footnotesize{MD} \\
    \midrule
  VGG19(BN)\cite{simonyan2014very} &70.35 &72.45 & \textbf{73.40} &73.03 & \textbf{73.98}  \\
  ResNet18\cite{he2016deep} &68.12 &69.84 & \textbf{70.52} & 68.93 & \textbf{69.82} \\
  ResNet50\cite{he2016deep} &73.56 &75.24 & \textbf{76.11} &74.73 & \textbf{75.59}  \\

    \bottomrule
\end{tabu}
\end{table*}
To verify the effectiveness of our method on the large-scale dataset, we also conduct experiments on ImageNet. We evaluate our method with three widely used networks including VGG19\cite{simonyan2014very}, ResNet18\cite{he2016deep} and ResNet50\cite{he2016deep}. The results are presented in Table~\ref{tab:ImageNet}. We observe that our method  consistently outperforms the self-distillation baseline -- our method improves the baseline by 0.95\%, 0.68\% and 0.87\% on VGG19, ResNet18 and ResNet50, respectively. These results clearly demonstrate the effectiveness of our method. More notably, our method can significantly improve the original models without introducing extra computational cost. The enhaced VGG19, ResNet18 and ResNet50 outperform the original models by 3.05\%, 2.40\% and 2.03\%, respectively. Since our method does not requires modifications on the original models during inference, MetaDistiller is very useful in boosting various prevalent CNN models for real-world applications.

\subsection{Comparison With Traditional Distillation}
\begin{table*}[!tbp]
\centering
\caption{\small \textbf{Our method outperforms the traditional distillation methods} We present the results on CIFAR-100 by testing our MetaDistiller with 5 two-stage distillation methods.}
\label{tab:distillation} 
\centering
    \begin{tabu}to\textwidth{l*{8}{X[c]}}\\\toprule
    {Teacher} & {Student} &  Baseline & KD~\cite{hinton2015distilling} & FitNet~\cite{romero2014fitnets} & AT~\cite{zagoruyko2016paying} & DML~\cite{zhang2018deep} & MD \\
    
    \midrule
  ResNet152\cite{he2016deep} & ResNet18\cite{he2016deep} &77.31 & 77.98 &78.41 & 78.81 & 77.72 & \textbf{79.05}\\
  ResNet152\cite{he2016deep} & ResNet50\cite{he2016deep} &77.78 & 79.39 & 80.20 & 79.47 & 78.45 & \textbf{80.85}  \\
    
    \bottomrule
\end{tabu}
\end{table*}
We further compare our method with some widely-used traditional knowledge distillation methods, including KD~\cite{hinton2015distilling}, FitNet~\cite{romero2014fitnets}, AT~\cite{zagoruyko2016paying} and DML~\cite{zhang2018deep}. All these methods follow the teacher-student pipeline. Concretely, they first train a ResNet of 152 layers (ResNet152) and then use the output from this large network as teacher to train a rather smaller network like ResNet18 and ResNet50. The experiments are conducted on CIFAR-100 dataset. We present the results in Table~\ref{tab:distillation}

Our method can consistently improve the directly trained model by even larger margin compared to these five traditional distillation methods even though we do not have a large network to serve as a teacher. We outperform the runner-up by $0.44\%$ and $0.78\%$ on ResNet18 and ResNet50 respectively. This shows that our approach improves not only the training speed but also the accuracy in comparison with traditional distillation methods. We assume this is due to the output from large teacher network cannot adapt well to the small model. This also proves the necessity to solve the capacity discrepancy problem in knowledge distillation.

\subsection{Ablation Study}
\begin{table*}[!tbp]
\centering
\caption{\small \textbf{Our method improves upon backbone by much larger margin compared to self-distillation.} We test on CIFAR-100 for ResNet of three different depth }
\label{tab:ablation} 
\centering
    \begin{tabu}to\textwidth{l*{3}{X[c]}}\\\toprule
    {Model} & {Method} &  {Final Output} & {Ensemble Output} \\
    \midrule
    \multirow{3}{*}{ResNet18~\cite{he2016deep}} & DSN\scriptsize{(Hard Only)}  & 78.01 & 79.18\\
    & SD\scriptsize{(Hard+Simple Soft)} & 78.25 & 79.32\\
    & MD\scriptsize{(Hard+Better Soft)} & \textbf{79.05} & \textbf{80.05}\\
    \midrule
    \multirow{3}{*}{ResNet50~\cite{he2016deep}} & DSN\scriptsize{(Hard Only)}  & 79.33 & 80.04\\
    & SD\scriptsize{(Hard+Simple Soft)} & 79.71 & 80.38 \\
    & MD\scriptsize{(Hard+Better Soft)} & \textbf{80.85} & \textbf{81.41}\\
    \midrule
    \multirow{3}{*}{ResNet101~\cite{he2016deep}} & DSN\scriptsize{(Hard Only)}  & 80.54 & 81.29 \\
    & SD\scriptsize{(Hard+Simple Soft)} & 80.92 & 81.55 \\
    & MD\scriptsize{(Hard+Better Soft)} & \textbf{81.39}& \textbf{81.98}\\

    \bottomrule
\end{tabu}
\end{table*}
There are major objectives for ablation study are two-folds. One is to show the improvement brought by our generated soft teacher targets is \textbf{much more} compared to self-distillation~\cite{zhang2019your} upon the deeply-supervised network (DSN), which only uses the hard label to supervise the intermediate outputs; and the other is to show our generated soft target and hard label are complementary, i.e. they both contribute to the training. We denote DSN, self-distillation and our approach as `Hard Only', `Hard+Simple Soft' and `Hard+Better Soft' respectively. We present all the results in Table.~\ref{tab:ablation}. We conduct the experiments for three ResNet-type networks on CIFAR-100.

From Table~\ref{tab:ablation}, we can observe that both hard label and soft targets contribute to the improvement of accuracy. Yet, the simple soft teacher target usually only improves upon DSN by around $0.4\%$, which is relatively smaller compared to using our generated soft target labels.  On the other hand, our method can improve the DSN by large margin, always more than $1\%$. This also indicates that our generated label is complementary to hard label given they can both contribute to boost the training, implying our generated soft teacher targets contain some helpful knowledge beyond the scope of ground-truth one-hot label. Notably, our generated soft targets bring more improvement than one-hot ground-truth label as well, which can be assumed that these hidden information is more helpful to the training of early stages compared to injecting more categorical information. These observations strongly demonstrate the effectiveness of the generated soft target, illustrating the benefits of using it to replace the final output as teacher label.

\subsection{Visualization and Discussion}

\begin{figure}[t]
  \centering
  \includegraphics[width=0.82\textwidth]{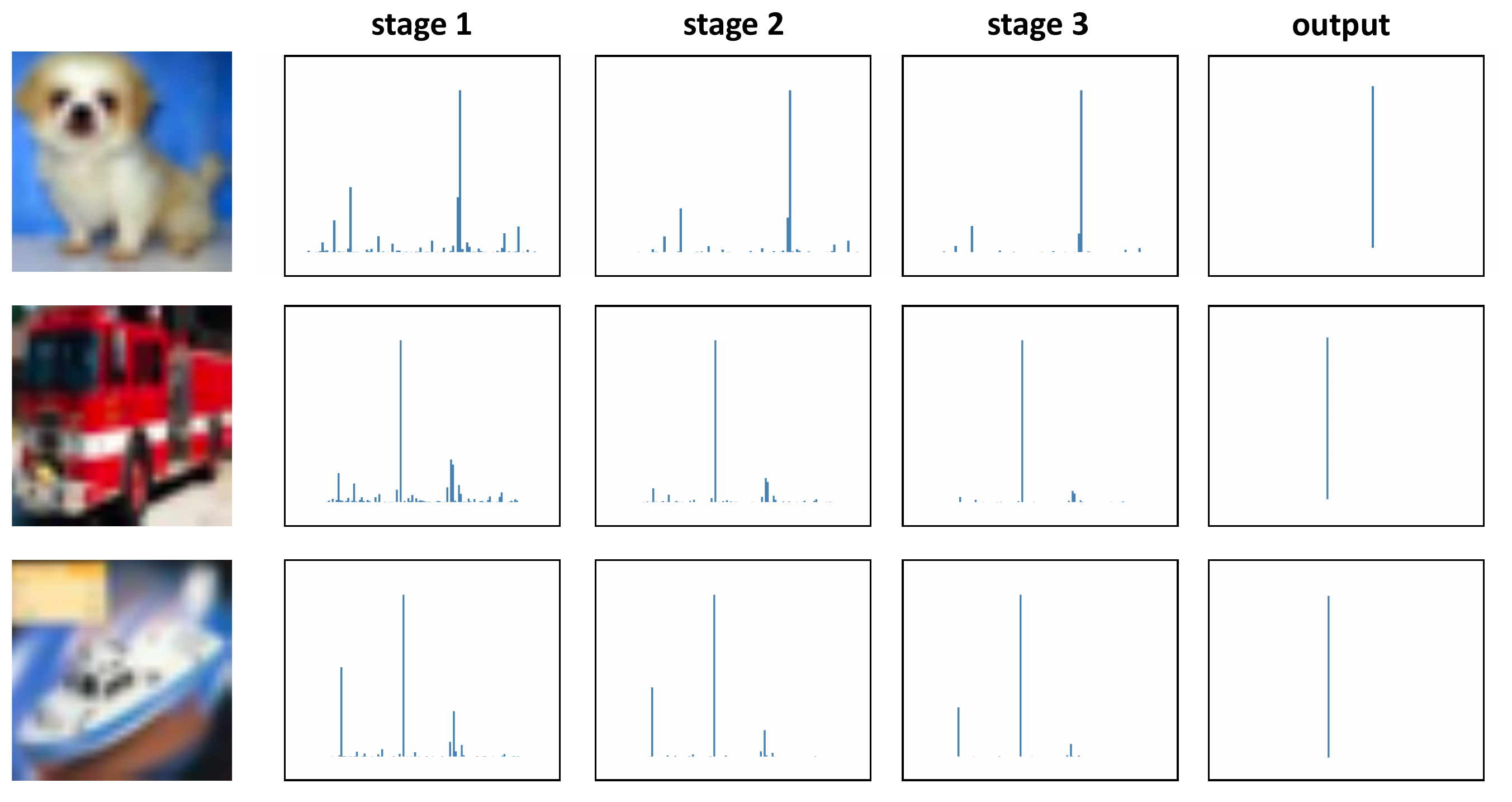}
  \caption{\small \textbf{Visualization.} We exhibit the output from a well-optimized ResNet18 and the generated soft targets of three intermediate stages produced by its corresponding label generator. We can see that the final output is quite similar to the one-hot distribution and the soft targets differs from each other and the ones of shallower stages is softer.}
  \label{fig:example}
\end{figure}

Why the generated soft teacher targets attain better results than using the final output? What `dark knowledge' does the generated soft teacher target contain but final output does not? To answer this question, we visualize the normalized generated soft targets for each intermediate stage as well as the final output categorical distribution both from ResNet18 in Fig.~\ref{fig:example}. We can find that the generated soft targets are smoother while the final output often approaches the one-hot ground-truth label. Due to this fact, the final output cannot provide enough information for training that are not covered by the hard label. Moreover, we can see that the generated targets for shallower layers are even softer than that for deep layers. We argue this is because the feature maps in shallow layers often encode more generic information while the deep layers contain more specific information. Therefore, using the output from deepest layer often lead to the compatibility problem~\cite{cho2019efficacy}. These observations imply that knowledge distillation is more like a kind of regularization, which avoids the model overfitting to the one-hot label on training data so that it can generalize well to unseen data.

The above phenomena are also observed by previous works~\cite{hinton2015distilling,cho2019efficacy,yang2018knowledge} and they solve this by fine-tuning the temperature~\cite{hinton2015distilling}, early stopping the teacher training~\cite{cho2019efficacy} and adding a regularization to the teacher to encourage it to be less strict~\cite{yang2018knowledge}. Our method achieves the same to make the teacher label more tolerant yet through a fully automatic way. We do not need to carefully fine-tune the hyper-parameter like temperature given our generated teacher label can learn to adapt itself to be more compatible. 

\section{Conclusion}

We proposed a new knowledge distillation method, namely MetaDistiller, to sidestep the time-consuming sequential training and improves the capability of the model at the same time. To achieve this, we generate more compatible soft teacher targets for the intermediate output layers by using a label generator to fuse the feature maps from deep layers of the main model in a top-down way. 
The experimental results on CIFAR and ImageNet prove the effectiveness of our method and the generalizability to different network architectures.

\noindent {\textbf{Acknowledgement. } Cho-Jui Hsieh would like to acknowledge the support by NSF IIS-1719097, Intel, and Facebook. This work was also supported by the National Key Research and Development Program of China under Grant 2017YFA-0700802, National
Natural Science Foundation of China under Grant 61822603, Grant U1813218,
Grant U1713214, and Grant 61672306, Beijing Natural Science Foundation
under Grant L172051, Beijing Academy of Artificial Intelligence, a grant from the Institute for Guo Qiang, Tsinghua
University, the Shenzhen Fundamental Research Fund (Subject Arrangement)
under Grant JCYJ20170412170-602564, and Tsinghua
University Initiative Scientific Research Program.}

%
%
\bibliographystyle{splncs04}
\bibliography{egbib}
\end{document}